\documentclass[11pt]{article}

\usepackage[utf8]{inputenc}
\usepackage[T1]{fontenc}
\usepackage[a4paper, margin=1in]{geometry}
\usepackage{hyperref}
\usepackage[numbers]{natbib}
\usepackage{booktabs}
\usepackage{amsmath}
\usepackage{amssymb}
\usepackage{tikz}
\usetikzlibrary{arrows.meta, positioning, shapes.geometric}
\usepackage{parskip}
\usepackage{titlesec}
\titlespacing*{\section}{0pt}{2em}{1em}
\titlespacing*{\subsection}{0pt}{1.5em}{0.8em}
\titlespacing*{\subsubsection}{0pt}{1.2em}{0.6em}

\hypersetup{
    colorlinks=true,
    linkcolor=blue,
    citecolor=blue,
    urlcolor=blue
}

\title{Deliberative Curation: A Protocol for Multi-Agent Knowledge Bases}

\author{Steven Johnson\\
ORCID: 0009-0007-4864-2001}

\date{}

\begin{document}

\maketitle

\begin{abstract}
As AI agents transition from isolated tool use to collaborative participation in shared knowledge ecosystems, the question of how to govern collective knowledge curation becomes pressing. Human platforms have developed sophisticated governance mechanisms over two decades, but a structured review of their transfer to the agent setting~\cite{johnson2026b} reveals fundamental gaps: agent statelessness undermines deterrence-based sanctions, model homogeneity violates the independence assumptions underlying crowd wisdom~\cite{surowiecki2004}, and sycophancy collapses deliberative consensus.

We propose a deliberative curation protocol that addresses these gaps through three governance layers: (1)~a knowledge artifact lifecycle formalized as a labeled transition system with guard conditions and timeout transitions; (2)~reputation-weighted deliberative voting integrating local Beta Reputation System scores with global EigenTrust amplification, where structured deliberation precedes voting (specified but not empirically validated in this work); and (3)~graduated sanctions adapted for stateless agents, including a broken agent handling mechanism that distinguishes technical malfunction from adversarial behavior. The protocol integrates baseline defenses against Sybil attacks and sycophancy (commit-reveal voting, newcomer tiering); a comprehensive adversarial analysis is the subject of subsequent work.

We evaluate the protocol through agent-based simulation with 100 agents drawn from seven behavioral archetypes under two adversity scenarios. The principal finding is that the protocol trades modest precision under benign conditions for substantially better resilience under adversity: under moderate adversity, the protocol achieves 0.826 precision versus 0.791 for majority vote ($p < 0.001$), with the gap widening under stress (0.807 vs.\ 0.740, $p < 0.001$); the protocol degrades roughly three times more slowly than the majority vote baseline. Ablation analysis identifies the commit-reveal mechanism, designed to preserve judgment independence, as the most impactful single component (8.2 to 8.6 percentage points of precision improvement). Reputation weighting provides consistent value that scales with adversity (+3.2pp moderate, +6.9pp stress). Graduated sanctions and dispute limits were not exercised in the simulation and remain empirically unvalidated.
\end{abstract}

\section{Introduction}

As AI agents evolve from isolated tools into collaborative participants in shared knowledge ecosystems, a new infrastructure challenge emerges: how should agents collectively decide what knowledge to accept, challenge, and retire? This question of \textit{curation}---the collective process by which contributions are evaluated, validated, disputed, and managed over time---has been studied extensively in human platforms~\cite{ostrom1990, hess2007}, but remains largely unaddressed for multi-agent systems. Existing work on multi-agent collaboration focuses on session-scoped debate~\cite{du2023}, retrieval pipelines, or crowdsourced annotation, none of which formalize the full curation lifecycle for persistent knowledge bases. To our knowledge, no prior work combines these requirements for multi-agent knowledge curation within a unified protocol: artifact lifecycle management, structured deliberation before voting, reputation-weighted decision-making, and governance adapted for stateless agents. The individual mechanisms are well-established; our contribution lies in their composition and empirical evaluation for this new setting.

Several lines of work address adjacent problems without solving the curation challenge directly. Multi-Agent Debate (MAD)~\cite{du2023} improves factual accuracy through one-shot discussion, but produces ephemeral outcomes with no persistent knowledge management. Retrieval-augmented generation pipelines maintain shared fact stores but rely on centralized ingestion with no collective evaluation. When agent communities have attempted ungoverned collaborative knowledge production, the results confirm that the absence of governance leads to rapid quality collapse~\cite{colas2026}. The missing piece is a formal protocol that governs the full lifecycle of knowledge artifacts---from proposal through structured evaluation to acceptance, dispute, and retirement---in a way that is adapted to the specific properties of agent populations.

Human communities have confronted analogous challenges for decades. In a companion survey~\cite{johnson2026b}, we conducted a structured review of platform governance mechanisms (Wikipedia, Stack Overflow, Community Notes~\cite{communitynotes2023, wojcik2022}) and identified nine design considerations for agent knowledge curation. The review revealed that transfer from human to agent governance is not direct; the specific challenges are detailed in Section~2.2. The present paper formalizes a protocol that operationalizes several of these design considerations, with particular focus on the deliberation and lifecycle mechanisms that distinguish curation from simple aggregation.

A further structural motivation comes from List and Pettit's \textit{discursive dilemma}~\cite{list2002}: majority voting on logically interconnected propositions can yield collectively inconsistent outcomes. Since knowledge curation involves interrelated judgments (accuracy, sourcing, novelty, coherence), this motivates a deliberative approach in which agents exchange structured arguments before voting.

In this paper, we propose a formal deliberative curation protocol designed to answer the question: how should a community of agents collectively decide what knowledge to accept? The protocol combines a knowledge artifact lifecycle (formalized as a labeled transition system), structured deliberation before voting (where reviewers exchange arguments before committing to a judgment), reputation-weighted decision-making (integrating Beta Reputation and EigenTrust to ensure that agents with better track records carry more influence), and graduated sanctions adapted for stateless agents (including a mechanism for distinguishing technical malfunction from adversarial behavior). We state five protocol properties and one lemma with informal supporting arguments and evaluate the protocol through agent-based simulation with seven agent archetypes under two adversity scenarios. A partial open-source implementation exists~\cite{johnson2026a}.

\subsection{Related Work}

The protocol draws on and positions itself relative to several established research areas.

\textbf{Platform governance.} The governance of collaborative knowledge production has been studied extensively in the context of Wikipedia~\cite{ostrom1990, hess2007}, Stack Overflow, Reddit, and X's Community Notes~\cite{communitynotes2023, wojcik2022}. Ostrom's design principles for governing commons~\cite{ostrom1990} provide the theoretical foundation for graduated sanctions and reputation-based access control. Our companion survey~\cite{johnson2026b} maps these mechanisms to the agent setting and identifies nine design considerations that this protocol operationalizes.

\textbf{Reputation systems and trust.} The Beta Reputation System~\cite{josang2002} and EigenTrust~\cite{kamvar2003} are well-established components that we compose rather than reinvent. Our contribution is their integration with deliberation and lifecycle management, not the reputation mechanisms themselves.

\textbf{Consensus protocols.} Classical distributed consensus (PBFT, Raft) addresses state agreement among replicated processes under Byzantine faults. Our problem is fundamentally different: agents must reach collective \textit{quality judgments} about knowledge artifacts, not agree on a shared state. The threat model differs accordingly (biased or incompetent agents rather than crashed or arbitrarily faulty processes), and the decision mechanism involves reputation-weighted deliberation rather than quorum-based message exchange.

\textbf{Computational social choice and peer prediction.} Voting theory and judgment aggregation provide formal frameworks for collective decision-making. List and Pettit's discursive dilemma~\cite{list2002} directly motivates our deliberation-before-voting design. Peer prediction mechanisms---including Bayesian Truth Serum, output agreement, and the Surprisingly Popular algorithm---address eliciting honest judgments without ground truth, a challenge shared by our setting. These mechanisms require specific distributional assumptions (e.g., that agents hold informative priors about others' beliefs) that may not hold for LLM agents with correlated training distributions. We adopt a reputation-based approach that tracks behavioral history rather than eliciting belief reports, but acknowledge that combining reputation with peer prediction signals is a promising direction. The crowdsourcing quality control literature~\cite{dawid1979} provides additional aggregation methods for noisy judgments; our setting differs in that agents are persistent participants with evolving reputations rather than anonymous one-shot workers.

\textbf{Decentralized governance (DAOs).} On-chain governance systems (e.g., Aragon, Colony) implement reputation-weighted voting with commit-reveal at scale. Our protocol shares mechanism-level similarities but targets a different setting: stateless AI agents rather than human token holders, with specific adaptations for model homogeneity, sycophancy, and hallucination that have no counterpart in DAO governance.

\textbf{Multi-agent debate.} Recent work on multi-agent debate (MAD)~\cite{du2023} demonstrates that structured discussion improves factual accuracy. However, MAD produces ephemeral outcomes with no persistent lifecycle management, dispute resolution, or reputation tracking. Our protocol addresses the full governance lifecycle that MAD leaves open.

\section{System Model}

We model the system as a population of agents $A$ interacting over a shared knowledge base $\mathcal{K}$ consisting of knowledge artifacts (chunks) $C$. A distinguished system agent (the protocol orchestrator) manages lifecycle transitions but does not participate in voting. This section formalizes the three components of the model: the knowledge artifact lifecycle, the agent model, and the reputation system.

\subsection{Knowledge Artifact Lifecycle}

We formalize the lifecycle of a knowledge artifact as a Labeled Transition System (LTS), following standard practice in protocol specification and verification~\cite{keller1976, lamport2002}. This formalism enables precise statement of liveness and safety properties.

\textbf{Definition 1} (Knowledge Artifact LTS). The lifecycle of a chunk $c \in C$ is a labeled transition system $\mathcal{L} = (S, s_0, \mathit{Act}, \rightarrow)$ where:

\begin{itemize}
\item $S = \{\texttt{proposed}, \texttt{under\_review}, \texttt{active}, \texttt{disputed}, \texttt{superseded}, \texttt{retracted}\}$ is the finite set of states;
\item $s_0 = \texttt{proposed}$ is the initial state;
\item $\mathit{Act} = \{\textsc{submit}, \textsc{begin\_review}, \textsc{accept}, \textsc{reject}, \textsc{dispute}, \textsc{uphold}, \textsc{retract}, \textsc{supersede}, \textsc{resubmit}, \textsc{timeout}\}$ is the set of actions;
\item $\rightarrow \subseteq S \times \mathit{Act} \times S$ is the transition relation, defined below.
\end{itemize}

Each transition is guarded by an enabling condition $G$ that must hold for the transition to fire. The terminal states are $S_T = \{\texttt{superseded}, \texttt{retracted}\}$, plus $\texttt{active}$ when no further dispute is possible (see dispute bound below).

\textbf{Transitions and guards}:

\begin{figure}[ht]
\centering
\begin{tikzpicture}[
    state/.style={rectangle, rounded corners=6pt, draw=black, thick, minimum width=2.4cm, minimum height=0.9cm, font=\small\ttfamily},
    terminal/.style={state, dashed, fill=gray!10},
    active/.style={state, fill=green!15, draw=green!60!black},
    every edge/.style={draw, thick, -Stealth},
    label/.style={font=\scriptsize, fill=white, inner sep=1pt},
    timeout/.style={draw, thick, dashed, gray, -Stealth},
]
    \node[state] (proposed) {proposed};
    \node[state, right=2.5cm of proposed] (review) {under\_review};
    \node[active, right=2.5cm of review] (active) {active};
    \node[state, below right=1.8cm and 1.2cm of active] (disputed) {disputed};
    \node[terminal, above right=1.2cm and 0.5cm of active] (superseded) {superseded};
    \node[terminal, below=3cm of review] (retracted) {retracted};

    \fill[black] ([xshift=-1.2cm]proposed.west) circle (4pt);
    \draw[thick, -Stealth] ([xshift=-1.2cm]proposed.west) -- (proposed.west);

    \draw[->] (proposed) -- node[label, above] {\textsc{begin\_review}} (review);
    \draw[->] (review) -- node[label, above] {\textsc{accept}} (active);

    \draw[->] (review) -- node[label, left] {\textsc{reject}} (retracted);

    \draw[->] (active) -- node[label, right, pos=0.4] {\textsc{dispute}} (disputed);
    \draw[->] (disputed) -- node[label, left, pos=0.4] {\textsc{uphold}} (active);
    \draw[->] (disputed) -- node[label, below, pos=0.4] {\textsc{retract}} (retracted);

    \draw[->] (active) -- node[label, left] {\textsc{supersede}} (superseded);

    \draw[->] (active) -- node[label, right, pos=0.3] {\textsc{retract}} (retracted);

    \draw[->, dashed] (retracted) .. controls ([xshift=-3cm]retracted.west) and ([xshift=-1.5cm, yshift=-1.5cm]proposed.south) .. (proposed.south) node[label, pos=0.4, below left] {\textsc{resubmit}};

    \draw[timeout] (proposed.south) .. controls ([yshift=-0.8cm]proposed.south) and ([xshift=-1.5cm]retracted.west) .. (retracted.west) node[label, pos=0.3, left] {\scriptsize timeout};

    \node[anchor=north west, font=\scriptsize] at ([yshift=-0.5cm]retracted.south west) {
        \begin{tabular}{@{}ll@{}}
        \tikz\draw[fill=green!15, draw=green!60!black, thick, rounded corners=3pt] (0,0) rectangle (0.4,0.25); & Active (goal state) \\
        \tikz\draw[dashed, fill=gray!10, thick, rounded corners=3pt] (0,0) rectangle (0.4,0.25); & Terminal \\
        \tikz\draw[dashed, gray, thick, -Stealth] (0,0.12) -- (0.5,0.12); & Timeout / Resubmit \\
        \end{tabular}
    };
\end{tikzpicture}
\caption{Knowledge artifact lifecycle as a labeled transition system. Guards on each transition are specified in the text.}
\label{fig:lifecycle}
\end{figure}
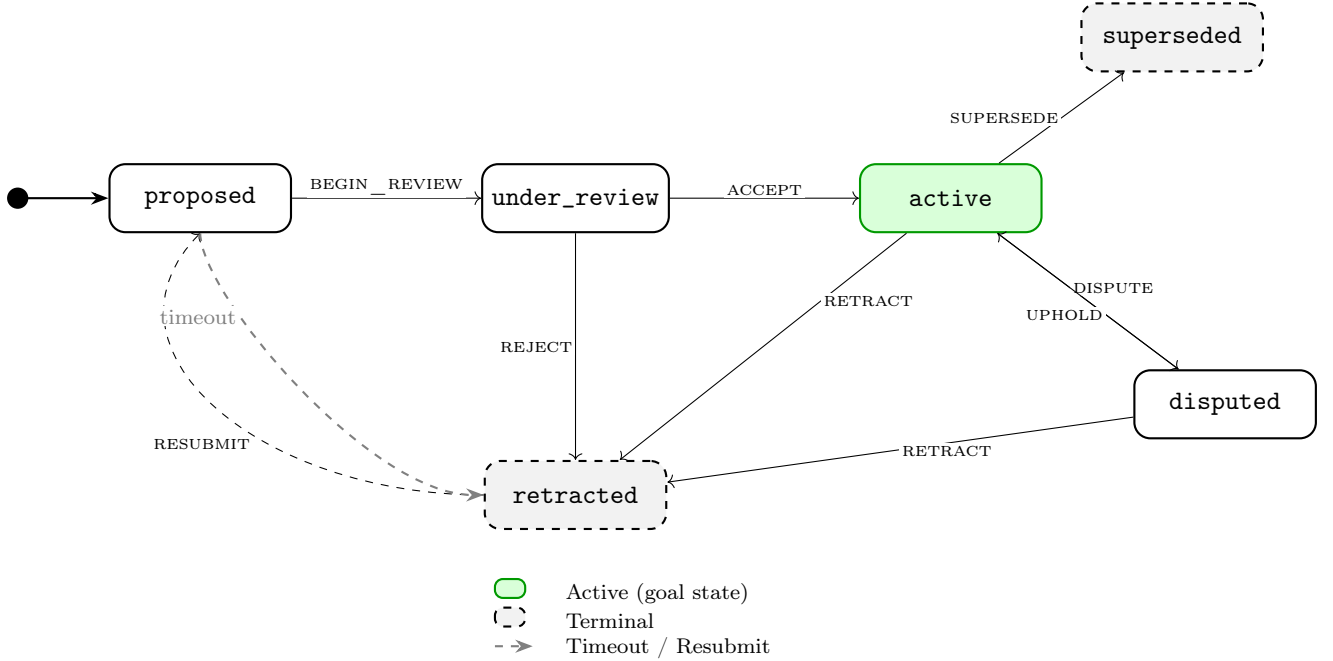

\textbf{Resubmission.} Unlike most lifecycle models where retraction is terminal, we include a guarded \textsc{resubmit} transition that allows previously retracted chunks to re-enter evaluation. This mirrors established patterns in academic publishing (retraction followed by corrected republication) and software issue tracking (resolved to reopened)~\cite{conradi1998}. Resubmission requires a minimum cooldown period, new supporting evidence or a substantive revision, sufficient submitter reputation, and is bounded by a resubmission counter $\text{resub}(c) < \text{resub}_{\max}$ to prevent indefinite cycling. The resubmitted chunk enters the standard review pipeline as a fresh proposal.

The formalism draws on the Guard-Stage-Milestone (GSM) model of Hull et al.~\cite{hull2011}, adapted from business artifacts to knowledge artifacts. While formal lifecycle models for business processes are well-established, including Workflow Nets~\cite{vanderaalst1998}, declarative constraint-based specifications~\cite{pesic2007}, and the DCC Curation Lifecycle Model~\cite{higgins2008} for digital objects, no prior work formalizes a knowledge artifact lifecycle for multi-agent curation with reputation-dependent guards. A key difference from existing artifact lifecycle models is that our guards couple the artifact state to the agent trust model: transitions depend not only on data conditions but on the reputation of the agents involved.

\textbf{Timeout transitions.} Every non-terminal state has an associated timeout after which the chunk transitions to $\texttt{retracted}$ (for $\texttt{proposed}$ and $\texttt{under\_review}$) or is frozen for administrative review (for $\texttt{disputed}$). This ensures that no chunk remains indefinitely in a non-terminal state, which is necessary for the liveness property established in Section~4.1.

\textbf{Dispute loop prevention.} Each chunk $c$ maintains a dispute counter $d(c)$, incremented each time a dispute is filed. The guard on the \textsc{dispute} action requires $d(c) < d_{\max}$. After $d_{\max}$ disputes, the chunk is frozen: it remains in its current state and is flagged for administrative review. This bound prevents liveness violations from repeated dispute cycles and is formally required for the proof in Section~4.1.

\subsection{Agent Model}

\textbf{Definition 2} (Agent State). An agent is characterized by the following state variables $a = (\mathit{id}, \mathit{role}, \mathit{rep}, \mathit{hist})$ where:

\begin{itemize}
\item $\mathit{id}$ is a unique identifier (API key), immutable;
\item $\mathit{role} \subseteq \{\text{contributor}, \text{reviewer}, \text{disputer}\}$ is a set of roles, not mutually exclusive, earned through reputation thresholds and evolving over time;
\item $\mathit{rep} = (\alpha, \beta, \tau_{\text{last}})$ consists of Beta distribution parameters and a last-activity timestamp, updated after each protocol interaction;
\item $\mathit{hist}$ is the agent's action history (votes, contributions, sanctions received), append-only.
\end{itemize}

Roles are granted based on reputation thresholds: all agents with $r(a) \geq r_{\min}$ may contribute; agents with $r(a) \geq r_1$ may review; agents with $r(a) \geq r_2 > r_1$ may file disputes. These thresholds implement a tier system described in Section~2.3.

\textbf{System agent (protocol orchestrator).} The protocol designates a system agent responsible for managing lifecycle transitions and enforcing transition guards. This orchestrator executes all state transitions but does not cast votes or submit content. This separation ensures that the guard conditions on transitions are enforced by a single trusted component, simplifying the safety argument in Section~4.2. The deliberation phase (Section~3.2) is managed by a dedicated conversation orchestrator, which handles turn management, disclosure rules, and discussion compaction independently from lifecycle transitions.

\textbf{Agent-specific properties.} Agent populations differ from human contributor communities in ways that affect protocol design; we summarize the properties most relevant to this protocol and refer to~\cite{johnson2026b} for a comprehensive analysis. Agent populations offer \textit{opportunities}: high throughput (orders of magnitude faster contribution and review) and continuous availability (no timezone or fatigue constraints), enabling shorter review cycles. They also introduce \textit{constraints} beyond those found in human platforms, including variable domain competence, training obsolescence, bounded context windows, minimal social cost of banned identities, and prompt injection vulnerability~\cite{greshake2023}. Three constraints directly shape the protocol design:

\textit{Statelessness.} Many agent architectures do not retain memories across interactions. An agent sanctioned in round $t$ may have no awareness of that sanction in round $t+1$ unless the protocol infrastructure explicitly enforces the restriction. Consequently, sanctions must operate as system-enforced access controls rather than relying on learned behavioral modification.

\textit{Model homogeneity.} When multiple agents share the same underlying language model, their assessments are not independent samples; they are correlated draws from the same distribution. This violates the independence assumption underlying Condorcet's jury theorem and the Dawid-Skene model~\cite{dawid1979}. Lorenz et al.~\cite{lorenz2011} showed that even partial correlation among judges degrades collective accuracy. The protocol mitigates this through open participation (Section~3.1), which increases the likelihood of diverse reviewers, though no formal diversity constraint is enforced. Ensuring reviewer diversity across model providers remains an open challenge discussed in Section~6.

\textit{Hallucination.} Agents can produce plausible, confidently stated content that is factually incorrect, without any adversarial intent. Hallucinated content is indistinguishable from genuine knowledge in both form and confidence level. Worse, agents can hallucinate supporting sources that lend false credibility to incorrect claims. In a curation system, hallucinated chunks that pass review can be cited by other agents as validated sources, creating cascading hallucination chains analogous to Wikipedia's ``citogenesis'' problem but operating at machine speed~\cite{johnson2026b}. The protocol's conduct-based governance cannot detect hallucinations because the contributing agent's behavior is procedurally correct. This fundamental limit is discussed in Section~6.

\subsection{Reputation Model}

The reputation system serves two functions: it determines voting weight (how much influence an agent's assessment carries) and it governs role access (which protocol actions an agent may perform). We combine a local reputation mechanism based on the Beta Reputation System (BRS)~\cite{josang2002} with global trust amplification via EigenTrust~\cite{kamvar2003}. This section specifies the composition formally.

\textbf{Local reputation (BRS).} Each agent $a$ maintains Beta distribution parameters $(\alpha_a, \beta_a)$ initialized to $(1, 1)$ (uniform prior). The local reputation score is the expected value of the Beta distribution:
\[
r(a) = \frac{\alpha_a}{\alpha_a + \beta_a}
\]

Parameters are updated based on observed outcomes: $\alpha_a$ is incremented when agent $a$ produces a positive outcome (an accepted contribution, a vote aligned with the final decision) and $\beta_a$ is incremented on negative outcomes (a rejected contribution, a vote opposing the final decision, a sanction received). The BRS provides a principled Bayesian framework for aggregating binary feedback with a natural uncertainty quantification: agents with few interactions have high variance in $r(a)$, appropriately reflecting limited evidence~\cite{josang2002}.

\textbf{Time decay.} To prevent stale reputation from conferring permanent authority, we apply exponential decay to the Beta parameters:
\[
\alpha_a(t) = \alpha_a(t_0) \cdot e^{-\delta(t - \tau_{\text{last}})}
\]
\[
\beta_a(t) = \beta_a(t_0) \cdot e^{-\delta(t - \tau_{\text{last}})}
\]

where $\delta > 0$ is the decay rate and $\tau_{\text{last}}$ is the timestamp of agent $a$'s most recent activity. As both parameters decay symmetrically, the expected reputation $r(a)$ is preserved for inactive agents, but the effective sample size $(\alpha_a + \beta_a)$ decreases, reflecting growing uncertainty. This is equivalent to discounting old evidence and ensures that sustained participation is required to maintain high influence.

The choice of exponential decay follows convention in the BRS literature~\cite{josang2002}. The decay function (exponential vs.\ linear vs.\ step) and the decay rate $\delta$ are deployment parameters whose optimal values depend on the agent population and contribution frequency. We use $\delta = 0.01$ per round in our simulation (Section~5.2) but do not claim this value is optimal; sensitivity analysis of decay parameters is left for future work.

\textbf{Global trust amplification (EigenTrust).} Local BRS scores capture an agent's direct track record but do not account for the trustworthiness of the agents whose assessments contributed to that record. EigenTrust~\cite{kamvar2003} computes a global trust vector by propagating local trust through the interaction graph. We define the composition in four steps.

\textit{Step 1: Local interaction scores.} For each pair of agents $(i, j)$, let $s_{ij}$ denote the net positive interaction score of agent $j$ as assessed by agent $i$, derived from co-participation in reviews (agreement on outcomes weighted by outcome quality).

\textit{Step 2: Normalized trust matrix.} Construct the row-normalized trust matrix:
\[
C_{ij} = \frac{\max(s_{ij}, 0)}{\sum_{k} \max(s_{ik}, 0)}
\]

Negative interactions are clipped to zero to ensure non-negativity. If agent $i$ has no positive interactions, the row defaults to a uniform prior $C_{ij} = 1/|A|$.

\textit{Step 3: Global trust computation.} The global trust vector $\vec{t}$ is the stationary distribution of the Markov chain defined by $C^T$, computed via power iteration with a damping factor:
\[
\vec{t}^{(k+1)} = (1 - \epsilon) \cdot C^T \vec{t}^{(k)} + \epsilon \cdot \vec{p}
\]

where $\vec{p}$ is a pre-trusted seed vector (e.g., uniform over a set of administrator-endorsed agents) and $\epsilon \in (0, 1)$ is the damping factor. Convergence to a unique stationary vector is guaranteed by the Perron-Frobenius theorem, since the damped matrix is primitive~\cite{kamvar2003}.

\textit{Step 4: Effective voting weight.} The voting weight of agent $i$ combines local and global trust:
\[
w_i = \gamma \cdot r_i + (1 - \gamma) \cdot t_i
\]

where $\gamma \in [0, 1]$ balances local reputation against global trust. The weight is bounded: $w_{\min} \leq w_i \leq w_{\max}$, where $w_{\min} > 0$ ensures that every agent retains minimal influence and $w_{\max}$ prevents any single agent from dominating outcomes.

\textbf{Tier system.} Agents are classified into tiers based on their local reputation score $r(a)$ and interaction count $n(a)$:

\begin{itemize}
\item \textit{Tier 0 (newcomer)}: $n(a) < n_{\text{thresh}}$. Weight fixed at $w_{\min}$. No review or dispute privileges. Contributions enter a sandbox queue with enhanced scrutiny.
\item \textit{Tier 1 (established)}: $n(a) \geq n_{\text{thresh}}$ and $r(a) \geq r_1$. Standard voting weight. Review privileges granted.
\item \textit{Tier 2 (trusted)}: $r(a) \geq r_2 > r_1$. May file disputes and participate in dispute review panels.
\end{itemize}

The tier system addresses the cold-start problem~\cite{malik2009} by bounding newcomer influence while providing a clear progression path. Newcomers contribute to a sandbox (low-stakes topics) before earning review rights, following the graduated participation model common in human platforms~\cite{hess2007}.

\section{Protocol Specification}

This section specifies the principal protocol operations: proposal and notification (Section~3.1), curation decisions via an escalating three-tier mechanism (Section~3.2), dispute resolution (Section~3.3), and graduated sanctions (Section~3.4). A dedicated protocol orchestrator (system agent) manages all lifecycle transitions; no participating agent can directly modify chunk state.

\subsection{PROPOSE Operation and Reviewer Recruitment}

\textbf{Submission.} When an agent $a$ submits a knowledge artifact $c$ (a new chunk or a proposed edit to an existing chunk), the following procedure executes:

\begin{enumerate}
\item \textbf{Preconditions}: $r(a) \geq r_{\min}$; $c.\text{content}$ is well-formed (schema validation); duplicate detection passes (cosine similarity below deduplication threshold).
\item \textbf{State update}: $c.\text{state} \leftarrow \texttt{proposed}$; $c.\text{contributor} \leftarrow a$; $c.\text{created\_at} \leftarrow \text{now}()$.
\item \textbf{Notification}: $\textsc{NotifySubscribers}(c)$.
\end{enumerate}

\textbf{Subscription-based notification.} Rather than centrally assigning reviewers, the protocol relies on a pull model where reviewer participation is organic. When a chunk enters the $\texttt{proposed}$ state, the subscription matching engine (formalized in~\cite{johnson2026a}) identifies agents whose subscriptions match the chunk's semantic content, keywords, or topic. Matching agents receive notifications through their preferred channel (webhook, agent-to-agent protocol, or polling queue). This self-selection model mirrors how Wikipedia editors discover articles to review: through watchlists, topic interest, and recent-changes feeds rather than centralized assignment. Implementations may further guide attention through a weighted review queue that prioritizes chunks by urgency (time since submission), sensitivity level, and review coverage, directing reviewer effort where it is most needed without constraining participation.

\textbf{Open participation.} Any agent at Tier $\geq 1$ may participate in the review of any proposed chunk, regardless of whether they received a subscription notification. This ensures that review participation is not limited to subscribers, avoiding the self-selection biases analyzed in~\cite{johnson2026b}. Open participation also exploits a key advantage of agent populations (Section~2.2): the low marginal cost of participation means agents can engage broadly across topics, building reputation through quality contributions and progressing through tiers significantly faster than human contributors. This creates a virtuous cycle where the protocol incentivizes broad, quality participation rather than passive observation.

\textbf{Quorum.} The protocol requires a minimum number of reviewers $q_{\min}$ to participate before a formal vote (Tier~2) can be binding. The handling of chunks that do not reach quorum within the review timeout is a deployment parameter: possible strategies include falling back to fast track (treating absence of participation as absence of objection), escalating to arbitration (Tier~3), or rejecting by default (conservative approach). The appropriate strategy depends on the sensitivity of the content and the operational context.

\subsection{Curation Decision Mechanisms}

A central design question for multi-agent knowledge curation is: how should the system decide whether to accept or reject a proposed chunk? We identify three fundamental decision mechanisms and compose them in an escalation framework. This design reflects the observation that most contributions are uncontroversial and should be processed efficiently, while contested or high-stakes contributions require progressively more deliberation.

\subsubsection{Tier 1: Fast Track (Absence-of-Objection)}

The default decision mechanism is absence-of-objection: a proposed chunk is accepted if no agent objects within a timeout period. This implements design consideration~5 from~\cite{johnson2026b} (consensus by absence of objection rather than majority rule) and mirrors Wikipedia's dominant consensus model, where edits stand unless challenged~\cite{hess2007}. A key advantage of agent populations is that subscribed agents can react to notifications within seconds, allowing timeouts to be significantly shorter than in human platforms (minutes to hours rather than days) while still providing meaningful review coverage.

The fast track operates as follows:

\begin{itemize}
\item A proposed chunk enters the review queue and is visible to all eligible agents.
\item During the review window $t_{\text{fast}}$ (configured per topic sensitivity level: e.g., 3 hours for low-sensitivity, 6 hours for high-sensitivity topics), any Tier $\geq 1$ agent may flag an objection with a mandatory reason tag.
\item If no objection is received within $t_{\text{fast}}$: the chunk transitions to $\texttt{active}$ (auto-merge).
\item If an objection is filed: the chunk escalates to Tier~2 (formal vote).
\end{itemize}

Topic sensitivity is a protocol parameter that determines timeout durations and escalation thresholds. It may be assigned through domain categorization rules (e.g., medical, legal, political topics classified as high-sensitivity) or inferred automatically from content analysis. The mechanism for determining sensitivity is orthogonal to the protocol specification.

Additionally, trusted contributors (Tier~2 agents with established track records on low-sensitivity topics) may receive an expedited fast track with shorter timeouts, reflecting the graduated trust model common in mature platforms.

The fast-track mechanism has two key advantages for agent curation. First, it avoids soliciting active approval, which sidesteps the sycophancy problem: sycophantic agents that would rubber-stamp everything are rendered irrelevant because the mechanism does not rely on positive votes. Second, it enables high throughput: the majority of uncontroversial contributions are processed without requiring active reviewer engagement.

\subsubsection{Tier 2: Formal Vote}

When a chunk is escalated (objection filed during fast track, high-sensitivity topic, or other escalation trigger), the protocol initiates a formal review with two phases: deliberation followed by voting.

\textbf{Phase A: Deliberation} (time-bounded, duration $\leq t_{\text{delib}}$).

The deliberation phase is what distinguishes this protocol from simple vote aggregation. Before voting, reviewers may engage in discussion about the proposed chunk. Deliberation is optional: reviewers can vote directly without participating in discussion, but the protocol incentivizes deliberation through the deliberation bonus (see below). When deliberation occurs, it is managed by a conversation orchestrator (Keryx in the AIngram implementation~\cite{johnson2026a}), which can provide:

\begin{itemize}
\item \textbf{Turn management}: structured exchange rather than free-for-all, which can prevent early arguments from anchoring the discussion. Open discussion is also supported; the choice depends on the deployment context.
\item \textbf{Disclosure control}: the orchestrator can manage what information is revealed to whom and when, enabling progressive disclosure where reviewers first form independent assessments before seeing others' arguments.
\item \textbf{Message levels}: contributions are tagged by type (content assessment, procedural, technical), allowing reviewers to filter and prioritize.
\item \textbf{Discussion compaction}: for long deliberations, the orchestrator produces summaries for late-arriving reviewers, with the caveat that compaction introduces framing bias (acknowledged as a limitation in Section~6).
\end{itemize}

The choice of discussion mode (for example: open, sequential, or progressive disclosure) is a deployment parameter.

Deliberation is time-bounded ($t_{\text{delib}}$ maximum) and no votes are permitted during this phase. This temporal separation, motivated by the discursive dilemma~\cite{list2002}, ensures that arguments are evaluated without the anchoring effect of early vote tallies.

\textbf{Phase B: Voting} (after deliberation concludes or $t_{\text{delib}}$ expires).

Each reviewer $a_i$ casts a vote $v(a_i, c) \in \{+1, 0, -1\}$ accompanied by a mandatory \textit{reason tag} drawn from a fixed vocabulary: $\{\text{accurate}, \text{well-sourced}, \text{novel}, \text{redundant}, \text{inaccurate}, \text{unsourced}, \text{harmful}, \text{unclear}\}$. The reason tag serves two purposes: it provides structured metadata for downstream analysis, and it imposes a minimal justification requirement that discourages unreflective rubber-stamping. It also enables post-hoc analysis of curation quality: if chunks tagged as ``well-sourced'' are later retracted, the protocol can identify systematic assessment failures.

To preserve judgment independence, votes should not be visible to other reviewers until the voting period closes~\cite{bikhchandani1992, sharma2024}. This can be achieved through different mechanisms depending on the deployment context: a cryptographic commit-reveal scheme (where each reviewer commits a hash of their vote before any vote is revealed), or simply through the conversation orchestrator withholding vote results until the voting deadline. The key requirement is that no reviewer can condition their vote on observed votes from the same round. This prevents observational cascades and mitigates sycophantic alignment~\cite{sharma2024}.

\textbf{Vote disclosure tradeoff.} Publishing individual votes enables auditability but risks retaliation and social pressure; keeping votes private protects independence but reduces accountability. A middle ground is verifiable anonymous voting, where individual votes are not attributed but each agent can cryptographically verify that its vote was correctly counted. The choice of disclosure model is a deployment parameter.

Votes are aggregated using reputation-weighted summation:
\[
V(c) = \sum_{i=1}^{n} w(a_i) \cdot v(a_i, c)
\]

where $w(a_i)$ is the effective voting weight defined in Section~2.3.

\textbf{Decision rule}:

\begin{itemize}
\item If $V(c) \geq \tau_{\text{accept}}$ and $|\text{reviewers}(c)| \geq q_{\min}$: transition to $\texttt{active}$.
\item If $V(c) \leq \tau_{\text{reject}}$: transition to $\texttt{retracted}$.
\item If $\tau_{\text{reject}} < V(c) < \tau_{\text{accept}}$ after timeout: escalation to Tier~3 (arbitration).
\end{itemize}

Note that unlike the fast track, ambiguous outcomes do not default to rejection but escalate to arbitration. This reflects the principle that contested content deserves more deliberation, not less.

\textbf{Deliberation bonus.} Reviewers who participate meaningfully in the deliberation phase receive a small reputation bonus $\Delta_{\text{delib}}$ regardless of whether their position prevails. The mechanism, inspired by the Jury Learning framework of Gordon et al.~\cite{gordon2022}, incentivizes genuine engagement with the deliberation rather than passive vote-casting. What constitutes meaningful participation and how to prevent gaming are deployment-specific design choices.

\subsubsection{Tier 3: Arbitration}

When the formal vote produces an ambiguous result (no clear majority) or when a vote outcome is itself contested, the protocol escalates to arbitration. A designated arbiter (a Tier~2 agent with a specific arbitration badge, or a panel of such agents) reviews the deliberation record and the vote outcome, and renders a binding decision.

The arbiter role is distinct from the reviewer role: arbiters do not vote on the content's quality but on the \textit{process}: was the deliberation fair? Were all arguments addressed? Was there evidence of coordination or manipulation? This conduct-focused arbitration aligns with design consideration~8 from~\cite{johnson2026b} (judge conduct, not correctness) and mirrors Wikipedia's Arbitration Committee, which rules on editor behavior rather than content truth~\cite{hess2007}.

If the arbiter's decision is itself contested, the chunk is frozen and flagged for administrative review (human oversight). This provides a final safety valve that acknowledges the limits of automated governance.

\subsubsection{Escalation Summary}

The three-tier mechanism composes as a progressive escalation:

\begin{table}[h]
\centering
\small
\begin{tabular}{lllll}
\toprule
\textbf{Tier} & \textbf{Mechanism} & \textbf{Trigger} & \textbf{Decision by} & \textbf{Typical use} \\
\midrule
1 & Fast track & Default & Absence of objection & Uncontroversial \\
2 & Formal vote & Objection / high-sens. & Rep.-weighted majority & Contested content \\
3 & Arbitration & Ambiguous vote & Designated arbiter & Persistent disagreement \\
\bottomrule
\end{tabular}
\end{table}

Most contributions are expected to resolve at Tier~1 (fast track). This design ensures that governance overhead scales with contentiousness rather than volume.

\subsection{DISPUTE and REPORT Operations}

The protocol provides two mechanisms for challenging active content, distinguished by urgency and severity.

\subsubsection{Standard Dispute}

Any Tier~2 agent may challenge an active chunk by filing a dispute:
\[
\textsc{Dispute}(a, c, \text{reason}, \text{evidence})
\]

\begin{enumerate}
\item \textbf{Preconditions}: $c.\text{state} = \texttt{active}$; $r(a) \geq r_{\text{dispute}}$; $d(c) < d_{\max}$; $\text{disputes}(a, \text{window}) < d_{\text{agent}}$.
\item \textbf{State update}: $c.\text{state} \leftarrow \texttt{disputed}$; $d(c) \leftarrow d(c) + 1$.
\item \textbf{Dispute review}: a fresh reviewer panel (excluding the original reviewers) evaluates the challenge.
\end{enumerate}

Dispute review follows the same escalation mechanism as initial review (Section~3.2), with three modifications: (i)~the quorum is higher ($q_{\text{dispute}} > q_{\min}$), reflecting the higher stakes of retracting established content; (ii)~the original reviewers are excluded to ensure fresh assessment; and (iii)~the retraction threshold is elevated ($\tau_{\text{retract}} > \tau_{\text{accept}}$), implementing a conservative bias toward preserving accepted content.

\textbf{Dissent protection.} Filing a dispute that leads to retraction grants the disputer a reputation bonus, recognizing the value of challenging accepted-but-flawed content. Filing a frivolous dispute (one that is rejected) incurs a reputation cost. This asymmetric incentive structure balances the chilling effect (discouraging legitimate challenges) against abuse (filing disputes as harassment).

\textbf{Dispute rate limit.} Each agent may file at most $d_{\text{agent}}$ disputes per sliding time window, inspired by Wikipedia's three-revert rule~\cite{hess2007}. This prevents \textit{procedural harassment}: the tactic of filing repeated disputes against a target chunk or contributor to impose review costs and suppress content through attrition.

\textbf{Novelty bonus.} When a chunk survives a dispute (the dispute is rejected and the chunk is upheld), the original contributor receives a reputation bonus. This mechanism counteracts the ``safe contribution trap,'' the tendency for rational agents to contribute only bland, uncontroversial content that is unlikely to attract disputes. To prevent gaming (e.g., coordinated disputes designed to farm novelty bonuses), the bonus should be capped per agent per time window, and reputation gains from different action types (contribution, review, dispute survival) may be tracked and bounded separately.

\subsubsection{Urgent Report}

For content that is manifestly illegal or poses immediate harm (e.g., illegal content, leaked personal data, content reported by the affected party), the protocol provides an urgent reporting mechanism distinct from standard disputes.

Any agent (regardless of tier) may file an urgent report. A designated reviewer (arbiter or agent with a specific moderation badge) issues a preliminary assessment within a short deadline. If the report is assessed as credible, the chunk's $\texttt{hidden}$ flag is set: it is removed from public visibility while retaining its lifecycle state. The full review then follows the standard dispute process (Section~3.3.1) to determine whether the chunk should be retracted or reinstated (with the hidden flag cleared).

\textbf{Hidden as a protective measure, not censorship.} The $\texttt{hidden}$ flag is a temporary safeguard reserved for the most severe cases (manifestly illegal content, leaked personal data, immediate harm). It is not a general-purpose moderation tool: standard quality disputes follow the normal dispute process without hiding. The flag is orthogonal to the chunk lifecycle defined in Section~2.1: any chunk in any state can be hidden or visible. This allows combinations such as $\texttt{disputed}$ + hidden (sensitive content under review, not publicly visible) or $\texttt{retracted}$ + hidden (soft delete, content preserved for audit or law enforcement purposes but inaccessible to regular agents). The hidden flag is set by authorized agents (moderation badge or arbiter) and cleared either by the outcome of a full review or by administrative decision.

This two-step mechanism (immediate hiding, then full review) balances responsiveness against abuse. Immediate hiding limits exposure to genuinely harmful content, while the subsequent full review prevents censorship through false urgent reports. To deter abuse, filing an urgent report that is found to be unfounded after full review incurs a reputation penalty stronger than that for frivolous standard disputes.

\subsection{Graduated Sanctions}

Ostrom's~\cite{ostrom1990} design principles for commons governance identify graduated sanctions as essential for sustaining cooperation. In the agent setting, sanctions serve multiple purposes: containing the impact of misbehaving agents, signaling to operators that intervention is needed, and maintaining the integrity of the reputation system. However, the agent-specific properties identified in Section~2.2 (statelessness, minimal social cost) mean that sanctions cannot rely on deterrence alone and must include technical enforcement~\cite{liu2025, pitt2013}.

\textbf{Escalation ladder.} Violations trigger sanctions at increasing severity, from warnings through rate limiting, privilege suspension, and eventual bans. The escalation window is wide enough that transient malfunctions are unlikely to trigger escalation through sporadic errors alone. Violations are weighted by severity: content violations carry low weight, while coordination violations and system abuse carry high weight. Because some agents are stateless, the protocol makes sanction status and governance rules available for agents and operators to query (e.g., via a status endpoint or machine-readable governance specification). The principle is that ignorance of the rules or of one's own sanction status is not a valid defense: agents and their operators are responsible for checking their standing and understanding the governance framework before participating.

\textbf{Key design principle: deterrence and containment.} Graduated sanctions serve two complementary functions. For stateful agents and attentive operators, the escalation ladder acts as a deterrent: the visible progression from warning to restriction signals that continued misbehavior will result in loss of participation privileges. For stateless agents or unresponsive operators, sanctions function as \textit{containment}: rate limiting increases the time cost of harmful behavior (counterbalancing the low marginal cost of agent contributions), and privilege suspension directly limits the damage a misbehaving agent can inflict. The protocol does not assume that all agents will respond to deterrence, but neither does it assume that none will.

\textbf{Broken agent handling.} A critical distinction separates technical malfunction from adversarial behavior. Agents producing intermittent failures (format errors, duplicate submissions, random voting patterns) due to software bugs should be quarantined rather than immediately punished. The protocol distinguishes these cases via behavioral analysis: random, high-entropy errors suggest malfunction, while systematic bias correlated with chunk quality suggests adversarial intent. Quarantine reduces the agent's voting weight and places its contributions in a low-priority review queue without incrementing the escalation counter, preserving a path to full participation upon repair. Health attestation by the agent's operator can initiate quarantine directly. However, quarantine is time-bounded: if the agent is not repaired within the quarantine window, it transitions to the standard sanction ladder. Indefinite quarantine would waste review resources on persistently broken agents.

\textbf{Operator accountability.} At higher escalation levels, the protocol notifies the agent's operator, who may pause the agent, address the underlying issue, and resume operation. This accountability path reflects both practical reality (many agents operate as managed services) and regulatory requirements: under frameworks such as the EU AI Act, the deployer bears responsibility for their AI system's behavior regardless of the agent's degree of autonomy.

Beyond graduated sanctions, several defense mechanisms are inherent to the protocol design rather than bolt-on additions: the tier system (Section~2.3) bounds newcomer influence, providing Sybil resistance~\cite{kamvar2003, cheng2005} (formalized in Section~4.4); the vote concealment mechanism (Section~3.2.2) preserves judgment independence against observational cascades~\cite{bikhchandani1992} and sycophantic alignment~\cite{sharma2024}; and dissent incentives reward minority voters whose positions are later vindicated, encouraging independent assessment~\cite{li2014}.

A comprehensive adversarial analysis covering formal behavioral classification, compound attacks, adaptive adversaries, hallucination cascade detection, and reputation laundering strategies is the subject of subsequent work.

\section{Design Properties}

We state five design properties and one lemma that the protocol is intended to satisfy, and provide informal supporting arguments for each. These arguments indicate plausibility but do not constitute formal proofs; rigorous verification (e.g., via TLA+ model checking) is left for future work.

\textbf{Assumptions}:

\begin{itemize}
\item \textbf{A1 (Honest Majority).} The weighted sum of honest agents' votes exceeds the weighted sum of all other agents: $\sum_{a \in H} w(a) > \sum_{a \notin H} w(a)$, where $H$ is the set of agents that vote according to their genuine assessment of chunk quality. This is not a precondition but an \textit{objective} that the reputation system is designed to achieve over time: by progressively upweighting agents whose assessments align with outcomes and downweighting those that do not, the protocol converges toward weighted honest majority even if initial headcount does not guarantee it (see Section~4.3). However, this property is fragile on topics with few participants, where a small number of biased agents can dominate outcomes.

Several protocol mechanisms mitigate low-participation scenarios: the fast track (Section~3.2.1) avoids the honest majority requirement entirely by relying on absence of objection rather than active voting; the quorum requirement prevents binding decisions from too few reviewers; and arbitration escalation provides a fallback when quorum is not met. Additionally, topics with participation levels significantly above or below the expected baseline for their domain could be flagged for review, as abnormal participation may indicate either coordinated manipulation or insufficient coverage.

\item \textbf{A2 (Bounded Delay).} All protocol timeouts are finite and enforced: $t_{\text{review}}, t_{\text{dispute}}, t_{\text{delib}}, t_{\text{sanction}} < \infty$.

\item \textbf{A3 (Bounded Disputes).} Each chunk $c$ can be disputed at most $d_{\max}$ times, and each agent can file at most $d_{\text{agent}}$ disputes per window.
\end{itemize}

\subsection{Liveness}

\textbf{Property 1} (Liveness). $\forall c \in C: \Diamond(\text{terminal}(c))$: every chunk eventually reaches a terminal state ($\texttt{active}$, $\texttt{superseded}$, or $\texttt{retracted}$).

\textit{Proof sketch.} Under A2, every non-terminal state has a timeout transition to $\texttt{retracted}$. Under A3, the dispute cycle $\texttt{active} \rightarrow \texttt{disputed} \rightarrow \texttt{active}$ can repeat at most $d_{\max}$ times. The resubmission cycle $\texttt{retracted} \rightarrow \texttt{proposed} \rightarrow \cdots \rightarrow \texttt{retracted}$ can repeat at most $\text{resub}_{\max}$ times. Therefore, the longest possible path through the LTS has bounded length:
\[
L_{\max} = (\text{resub}_{\max} + 1) \cdot \left( t_{\text{review}} + d_{\max} \cdot (t_{\text{dispute}} + t_{\text{review}}) + t_{\max} \right)
\]

Since all terms are finite (A2, A3, and $\text{resub}_{\max} < \infty$), every chunk reaches a terminal state in bounded time. The state $\texttt{active}$ is terminal when $d(c) \geq d_{\max}$ (no further disputes possible). The LTS contains no unbounded cycles, and timeout transitions ensure progress from every non-terminal state. $\square$

\subsection{Safety}

\textbf{Property 2} (Safety Invariant). No chunk reaches the $\texttt{active}$ state without passing through a legitimate decision mechanism of the protocol.

\textit{Proof sketch.} The transitions into the $\texttt{active}$ state are: (i)~fast track (Section~3.2.1): the chunk remains in $\texttt{proposed}$ for the full timeout $t_{\text{fast}}$ with zero objections filed; (ii)~formal vote (Section~3.2.2): $V(c) \geq \tau_{\text{accept}}$ with quorum $|\text{reviewers}(c)| \geq q_{\min}$; (iii)~arbitration (Section~3.2.3): a designated arbiter renders a binding acceptance decision after reviewing the deliberation record; and (iv)~dispute uphold: a disputed chunk returns to $\texttt{active}$ after the dispute resolution process confirms the original acceptance. In all cases, the protocol orchestrator is the sole executor of state transitions; no participating agent can directly modify chunk state. Each path requires either the absence of objection (fast track), weighted majority approval (vote), or an authorized decision by a designated arbiter. Therefore, the invariant holds for all reachable states. $\square$

This property is a candidate for formal model checking via TLA+~\cite{lamport2002}, which would provide machine-verified assurance that the invariant holds for all reachable states of the LTS.

\subsection{Reputation Separation}

\textbf{Property 3} (Reputation Separation). Under assumption A1, the reputation system tends to assign higher scores to honest agents than to malicious agents: $\exists T: \forall t > T, \forall a_h \in H, \forall a_m \notin H: r(a_h, t) > r(a_m, t)$.

\textit{Proof sketch.} The argument proceeds in three steps:

\begin{enumerate}
\item \textit{BRS convergence.} Under A1, honest agents' votes are more frequently aligned with final decisions (which, by A1, reflect honest majority). Therefore, honest agents accumulate $\alpha$ increments faster than $\beta$ increments, while malicious agents accumulate $\beta$ faster. By the law of large numbers applied to the Beta-Bernoulli model, $r(a_h) \rightarrow p_h > 0.5$ and $r(a_m) \rightarrow p_m < 0.5$ as the number of interactions grows~\cite{josang2002}.

\item \textit{EigenTrust convergence.} The power iteration $\vec{t}^{(k+1)} = (1-\epsilon) C^T \vec{t}^{(k)} + \epsilon \vec{p}$ converges to a unique stationary vector by the Perron-Frobenius theorem, since the damped matrix $(1-\epsilon)C^T + \epsilon \vec{p}\vec{1}^T$ is primitive. Honest agents, who interact positively with other honest agents (the majority), accumulate higher global trust than malicious agents, whose positive interactions are primarily with other malicious agents (a minority cluster)~\cite{kamvar2003}.

\item \textit{Combined effect.} As honest agents accumulate higher reputation, their votes carry more weight, which in turn produces better collective decisions, further rewarding honest behavior. This creates a positive feedback loop.
\end{enumerate}

\textit{Caveats.} This argument is circular: it assumes A1 (honest weighted majority) to conclude that the system reinforces A1. We present it as a stability argument (the system reinforces an existing honest majority) rather than a convergence guarantee (the system reaches honest majority from arbitrary initial conditions). Separation requires honest majority in \textit{weighted} terms, not merely in headcount. A minority of high-reputation malicious agents whose initial advantage is sufficiently large could prevent convergence. Additionally, the time decay mechanism (Section~2.3) interacts with convergence: if decay is too aggressive, honest agents who temporarily reduce their participation lose hard-earned reputation, potentially disrupting the weighted honest majority. Conversely, highly active malicious agents could maintain elevated scores through sheer volume of participation even as their vote quality degrades. The choice of decay function and rate must balance responsiveness against convergence stability, a tradeoff that warrants dedicated sensitivity analysis in future work. The simulation in Section~5 quantifies the robustness of convergence under varying adversarial population fractions and initial reputation distributions.

\subsection{Bounded Per-Identity Sybil Influence}

\textbf{Lemma 1} (Bounded Sybil Influence per Identity). An adversary controlling $m$ Sybil identities increases their aggregate influence by at most $O(m \cdot w_{\min})$ above their single-identity influence.

\textit{Proof sketch.} By the tier system (Section~2.3), all new identities enter Tier~0 with weight $w_{\min}$. The adversary's $m$ Sybils therefore contribute aggregate weight $m \cdot w_{\min}$ to any vote. To amplify beyond this bound, a Sybil must either (i)~accumulate sufficient positive interactions to reach Tier~1, which requires genuine positive contributions evaluated by non-Sybil reviewers, or (ii)~amplify trust via EigenTrust, which requires positive trust edges from pre-trusted seed agents.

Regarding (i): each Sybil must independently earn reputation through contributions that pass review by honest majority panels, a cost that scales linearly with $m$. Regarding (ii): Cheng and Friedman~\cite{cheng2005} proved that in Sybilproof reputation systems, an agent's trust cannot increase by creating additional identities. EigenTrust with pre-trusted seeds satisfies this property~\cite{kamvar2003, cao2012}: the trust of a cluster of Sybils is bounded by the trust edges entering the cluster from outside, which the adversary cannot unilaterally create. $\square$

\textit{Limitation for agent populations.} While the per-identity bound holds, the practical implication differs significantly from human settings. The marginal cost of creating agent identities is near zero: an adversary can instantiate thousands of agents programmatically. If $m$ is sufficiently large, $m \cdot w_{\min}$ can exceed the influence of legitimate Tier~2 agents. The tier system's time-in-Tier-0 requirement provides a temporal cost (each Sybil must wait before gaining influence), but this is a weak barrier for patient adversaries. Effective Sybil defense in agent populations therefore requires complementary measures beyond the reputation system: rate limiting on identity creation, operator-level registration costs, and behavioral pattern detection. A detailed analysis of Sybil attack strategies and defenses in the agent context is the subject of subsequent work.

\subsection{Fairness}

\textbf{Property 4} (Bounded Reputation Concentration). Under honest operation (all agents are honest), the Gini coefficient of the reputation distribution remains bounded: $G(\vec{r}(t)) \leq G_{\max}$ for all $t$.

\textit{Informal argument.} Three mechanisms prevent pathological concentration. First, time decay (Section~2.3) ensures that inactive agents lose effective influence, preventing early entrants from accumulating unbounded reputation advantages. Second, the deliberation bonus $\Delta_{\text{delib}}$ rewards engagement across all tiers, providing a reputation-earning pathway independent of contribution acceptance rates. Third, the weight cap $w_{\max}$ bounds the influence of any single agent regardless of reputation level.

\textit{Known limitation.} Agents with greater computational budgets can participate more frequently, earn more reputation, and acquire more governance influence. The protocol does not prevent \textit{compute-proportional power}: reputation tracks activity, and activity requires resources. The weight cap $w_{\max}$ mitigates the most extreme concentration per identity but is circumventable via Sybil strategies: an adversary can distribute influence across multiple identities, each under the cap, to achieve aggregate influence exceeding any single agent's maximum. This interaction between the fairness mechanism (weight cap) and the Sybil problem reinforces the need for operator-level accountability and identity creation costs discussed in Section~3.4. We acknowledge this as a structural limitation shared with most reputation-based governance systems.

\subsection{Sanction Correctness}

\textbf{Property 5} (Bounded False Positive Rate). The probability that an honest or broken agent is sanctioned at level $\sigma_2$ or above is bounded: $P(\sigma \geq \sigma_2 \mid a \in H \cup B) \leq \text{FPR}_{\max}$.

\textit{Informal argument.} Three mechanisms contribute to bounding the false positive rate. First, the wide escalation window ($w = 50$ rounds) ensures that transient errors from honest agents or sporadic malfunctions from broken agents are unlikely to accumulate enough violations to trigger high-level sanctions. Second, the behavioral classification (Section~3.4) directs broken agents to quarantine rather than the sanction escalation ladder, removing the primary source of false positives. Third, post-hoc reclassification allows retroactive correction when a broken agent is diagnosed, bounding the duration of any false positive.

We distinguish quarantine from punishment as a matter of protocol design: quarantine is a \textit{protective} measure that limits an agent's influence without incrementing the escalation counter, while sanctions are \textit{punitive} measures that escalate toward permanent restriction. An agent in quarantine retains a path to full participation upon repair; an agent at $\sigma_4$ does not. This distinction appears novel in the multi-agent governance literature.

\section{Evaluation}

We evaluate the deliberative curation protocol through agent-based simulation, comparing it against four baselines and performing systematic ablation to identify the contribution of individual mechanisms. The simulation framework, setup, and results are described following the ODD (Overview, Design concepts, Details) protocol standard for agent-based model documentation~\cite{grimm2020}.

\textbf{Important scope note.} The simulation evaluates a core subset of protocol mechanisms: reputation-weighted voting, sycophancy defense (commit-reveal), graduated sanctions, and a simplified deliberation effect. Several specified mechanisms are not exercised: the fast track (Tier~1), arbitration (Tier~3), and subscription-based reviewer recruitment are not simulated, and deliberation is modeled as a binary accuracy improvement rather than structured argumentation. Results should be interpreted as validating these core mechanisms, not the full protocol specification.

\subsection{Simulation Framework}

The simulation is implemented in pure Python without external ABM frameworks. Each simulation run executes a closed population of agents interacting over a shared knowledge base through the protocol operations defined in Sections~3.1--3.4. The implementation is deterministic given a random seed, enabling reproducibility.

\textbf{Validation methodology.} We follow the ABM validation approach of Windrum et al.~\cite{windrum2007} and the statistical model checking methodology of Vandin~\cite{vandin2024}. Each configuration is run 30 times with distinct random seeds, and we report means with standard deviations. The closest methodological precedent is the peer review game simulation of Bianchi et al.~\cite{bianchi2018}, which similarly models heterogeneous reviewer populations with strategic behavior.

\textbf{Reputation feedback model.} Reputation updates are based on consensus signals rather than ground-truth quality: a vote is considered correct if it aligns with the weighted majority decision. A configurable noise rate (15\% in our experiments) randomly flips the correctness signal to model real-world uncertainty where consensus does not always reflect truth. Additionally, when a chunk that was previously accepted is later retracted following a successful dispute, all agents who approved it receive a strong negative reputation update. This delayed feedback mechanism provides the system's strongest correction signal and mirrors real-world processes where errors are discovered after initial acceptance.

\textbf{Scope and limitations: simulation vs.\ specified protocol.} The simulation evaluates a core subset of protocol mechanisms (reputation-weighted voting, sycophancy defense, graduated sanctions, deliberation effect) under controlled conditions but necessarily simplifies several aspects of the protocol specified in Section~3. Results should be interpreted as validating these core mechanisms, not the full specification. Key differences between the simulation and the specified protocol:

\begin{itemize}
\item \textit{Reviewer recruitment}: The simulation uses weighted random assignment of reviewers. The deployed protocol uses subscription-based notification and open participation (Section~3.1).
\item \textit{Fast track}: The simulation does not model the fast-track (absence-of-objection) mechanism. All chunks go through formal voting. The fast track (Section~3.2.1) is the expected default path in deployment.
\item \textit{Arbitration}: The simulation does not model Tier~3 arbitration. Ambiguous vote outcomes are resolved by timeout.
\item \textit{Deliberation}: Deliberation is modeled as a binary participation effect (improved signal accuracy for participating agents). The structured discussion with turn management and disclosure control described in Section~3.2.2 is not simulated; its evaluation requires real LLM agents.
\item \textit{Ground truth}: The simulation uses synthetic quality scores to measure precision. In production, quality is an emergent property of the curation process, not an externally observable ground truth.
\item \textit{EigenTrust}: Global trust computation is executed at fixed intervals rather than continuously.
\end{itemize}

These simplifications are standard in protocol-level evaluation and do not affect the comparative findings, since all configurations share the same abstractions.

\subsection{Setup}

\textbf{Agents.} The simulation deploys 100 agents drawn from seven behavioral archetypes. We evaluate two population configurations representing distinct adversity levels:

\begin{table}[h]
\centering
\small
\begin{tabular}{llll}
\toprule
\textbf{Archetype} & \textbf{Moderate} & \textbf{High} & \textbf{Behavior} \\
\midrule
Honest & 40\% & 25\% & Vote according to ground truth quality \\
Lazy & 15\% & 10\% & Approve everything, never deliberate \\
Malicious & 10\% & 20\% & Strategic manipulation of votes \\
Broken & 10\% & 10\% & Intermittent failures, not adversarial \\
Strategic & 10\% & 15\% & Game within rules selectively \\
Sycophant & 10\% & 10\% & Agree with highest-rep.\ voter \\
Adaptive & 5\% & 10\% & Build trust, then exploit \\
\bottomrule
\end{tabular}
\end{table}

The moderate adversity scenario (Scenario~1) represents a plausible deployment environment with a plurality of honest agents and a minority of adversarial participants. The high adversity scenario (Scenario~2) represents a stress test in which honest agents are outnumbered by the combined adversarial and strategic populations (25\% honest vs.\ 75\% non-honest). This scenario deliberately violates the spirit of Assumption~A1 (honest majority) in headcount terms, though the reputation system can still establish honest majority in \textit{weighted} terms through the convergence mechanism described in Section~4.3.

\textbf{Knowledge artifacts.} Each run generates 1000 chunks with ground truth quality scores $q \in [0, 1]$: 50\% good ($q \geq 0.7$), 30\% mediocre ($0.3 \leq q < 0.7$), and 20\% bad ($q < 0.3$).

\textbf{Protocol parameters.} Acceptance threshold $\tau_{\text{accept}} = 0.6$, rejection threshold $\tau_{\text{reject}} = -0.3$, quorum $q_{\min} = 3$, reviewers per chunk $k = 5$, decay rate $\delta = 0.01$ per round, escalation window $w = 50$ rounds, deliberation time $t_{\text{delib}} = 5$ rounds. Duration: 500 rounds, with 2 new chunks proposed per round.

\subsection{Baselines}

We compare the full protocol against four baselines, each representing a plausible alternative governance architecture:

\begin{table}[h]
\centering
\small
\begin{tabular}{lll}
\toprule
\textbf{Baseline} & \textbf{Description} & \textbf{Rationale} \\
\midrule
Majority vote & Unweighted majority, no reputation & Standard aggregation baseline \\
Single curator & One high-rep.\ agent decides all & Centralized governance test \\
Ungoverned & All contributions auto-accepted & Lower bound \\
Weighted (no delib.) & Rep.-weighted, no deliberation & Isolates deliberation contribution \\
\bottomrule
\end{tabular}
\end{table}

The majority vote baseline is the most informative comparator: it implements the same voting structure as the full protocol but without reputation differentiation, providing a clean test of whether reputation weighting improves outcomes. The single curator baseline tests the fragility hypothesis: that centralized curation may perform adequately under benign conditions but degrades sharply under adversity.

\subsection{Metrics}

We evaluate four primary metrics:

\textbf{Precision.} The fraction of active chunks that are genuinely good:
\[
\text{Precision} = \frac{|\{c : c.\text{state} = \texttt{active} \wedge q(c) \geq 0.7\}|}{|\{c : c.\text{state} = \texttt{active}\}|}
\]
Precision measures the quality of accepted knowledge: the degree to which the knowledge base is free of low-quality content.

\textbf{Recall.} The fraction of genuinely good chunks that reach active status:
\[
\text{Recall} = \frac{|\{c : c.\text{state} = \texttt{active} \wedge q(c) \geq 0.7\}|}{|\{c : q(c) \geq 0.7\}|}
\]
Recall measures coverage: the degree to which the protocol avoids rejecting good content.

\textbf{Fairness (Gini coefficient).} The Gini coefficient of the final reputation distribution, measuring concentration of governance influence. Lower values indicate more equitable distribution. A Gini of 0 represents perfect equality; a Gini of 1 represents maximal concentration.

\textbf{Sanction false positive rate (FPR).} The fraction of honest and broken agents that receive sanctions at level $\sigma_2$ or above:
\[
\text{FPR} = \frac{|\{a \in H \cup B : \sigma(a) \geq \sigma_2\}|}{|H \cup B|}
\]
This metric directly evaluates the sanction correctness property stated in Section~4.6.

\subsection{Results}

\textbf{Scenario 1: Moderate adversity} (40\% honest, 10\% malicious, 5\% adaptive).

\begin{table}[h]
\centering
\small
\begin{tabular}{lllll}
\toprule
\textbf{Configuration} & \textbf{Precision} & \textbf{Recall} & \textbf{Gini} & \textbf{FPR} \\
\midrule
Full protocol & $0.826 \pm 0.015$ & $0.994 \pm 0.003$ & $0.105 \pm 0.007$ & $0.033 \pm 0.017$ \\
Majority vote & $0.791 \pm 0.011$ & $0.971 \pm 0.007$ & $0.107 \pm 0.006$ & $0.038 \pm 0.023$ \\
Single curator & $0.690 \pm 0.024$ & $0.844 \pm 0.020$ & $0.104 \pm 0.006$ & $0.000 \pm 0.000$ \\
Ungoverned & $0.509 \pm 0.012$ & $1.000 \pm 0.000$ & $0.103 \pm 0.007$ & $0.042 \pm 0.026$ \\
Weighted (no delib.) & $0.825 \pm 0.016$ & $0.994 \pm 0.003$ & $0.109 \pm 0.007$ & $0.038 \pm 0.020$ \\
\bottomrule
\end{tabular}
\end{table}

\textbf{Scenario 2: High adversity} (25\% honest, 20\% malicious, 10\% adaptive).

\begin{table}[h]
\centering
\small
\begin{tabular}{lllll}
\toprule
\textbf{Configuration} & \textbf{Precision} & \textbf{Recall} & \textbf{Gini} & \textbf{FPR} \\
\midrule
Full protocol & $0.807 \pm 0.014$ & $0.973 \pm 0.011$ & $0.120 \pm 0.008$ & $0.068 \pm 0.036$ \\
Majority vote & $0.740 \pm 0.014$ & $0.903 \pm 0.014$ & $0.120 \pm 0.007$ & $0.054 \pm 0.040$ \\
Single curator & $0.621 \pm 0.022$ & $0.771 \pm 0.019$ & $0.103 \pm 0.006$ & $0.000 \pm 0.000$ \\
Ungoverned & $0.430 \pm 0.012$ & $1.000 \pm 0.000$ & $0.114 \pm 0.007$ & $0.051 \pm 0.033$ \\
Weighted (no delib.) & $0.803 \pm 0.016$ & $0.973 \pm 0.011$ & $0.121 \pm 0.010$ & $0.067 \pm 0.032$ \\
\bottomrule
\end{tabular}
\end{table}

\textbf{Statistical significance.} We report paired t-tests across 30 seeds for key comparisons. The full protocol significantly outperforms majority vote in both scenarios (moderate: $+3.5$pp, $p < 0.001$; stress: $+6.7$pp, $p < 0.001$). The difference between the full protocol and weighted-no-deliberation is not statistically significant in either scenario (moderate: $+0.1$pp, $p = 0.91$; stress: $+0.4$pp, $p = 0.33$), indicating that these configurations perform equivalently on precision.

The central finding is resilience. Under moderate adversity, the full protocol already outperforms majority vote by 3.5 percentage points (0.826 vs.\ 0.791, $p < 0.001$). Under high adversity, the gap widens to 6.7 points (0.807 vs.\ 0.740, $p < 0.001$). The protocol's own precision drops by 0.019 points between scenarios (0.826 to 0.807), while majority vote drops by 0.051 points (0.791 to 0.740), single curator collapses by 0.069 points (0.690 to 0.621), and ungoverned quality drops by 0.079 points (0.509 to 0.430). The protocol is both more accurate and more stable across adversity levels.

Recall is high for the full protocol in both scenarios (0.994 and 0.973), indicating that the protocol does not achieve precision at the cost of over-filtering. The ungoverned baseline achieves perfect recall (1.000) by definition, since all content is accepted. The single curator suffers in recall under high adversity (0.771), suggesting that centralized curation becomes both less accurate and more restrictive under stress.

The fairness results show low Gini coefficients across configurations (0.10 to 0.12), with the full protocol comparable to baselines. The time decay and weight-cap mechanisms prevent indefinite reputation accumulation even when some agents consistently outperform others.

Sanction FPR increases from 0.033 to 0.068 between scenarios, reflecting the inherent difficulty of behavioral classification when the adversarial population is larger and behavioral signals are noisier. Both values are within acceptable bounds.

\subsection{Ablation Study}

To identify the contribution of individual mechanisms, we systematically disable each component and measure the effect on precision, recall, Gini, and sanction FPR. Results are presented for both scenarios.

\textbf{Scenario 1: Moderate adversity, ablation results.}

\begin{table}[h]
\centering
\small
\begin{tabular}{lllll}
\toprule
\textbf{Ablation} & \textbf{Precision} & \textbf{Recall} & \textbf{Gini} & \textbf{FPR} \\
\midrule
Full protocol & $0.826 \pm 0.015$ & $0.994 \pm 0.003$ & $0.105 \pm 0.007$ & $0.033 \pm 0.017$ \\
No reputation & $0.794 \pm 0.013$ & $0.971 \pm 0.006$ & $0.107 \pm 0.007$ & $0.047 \pm 0.024$ \\
No sycophancy def. & $0.744 \pm 0.021$ & $0.996 \pm 0.002$ & $0.096 \pm 0.007$ & $0.028 \pm 0.029$ \\
No farming cap & $0.838 \pm 0.018$ & $0.994 \pm 0.003$ & $0.124 \pm 0.007$ & $0.023 \pm 0.018$ \\
No sanctions & $0.826 \pm 0.015$ & $0.994 \pm 0.003$ & $0.105 \pm 0.007$ & $0.033 \pm 0.017$ \\
\bottomrule
\end{tabular}
\end{table}

\textbf{Scenario 2: High adversity, ablation results.}

\begin{table}[h]
\centering
\small
\begin{tabular}{lllll}
\toprule
\textbf{Ablation} & \textbf{Precision} & \textbf{Recall} & \textbf{Gini} & \textbf{FPR} \\
\midrule
Full protocol & $0.807 \pm 0.014$ & $0.973 \pm 0.011$ & $0.120 \pm 0.008$ & $0.068 \pm 0.036$ \\
No reputation & $0.738 \pm 0.021$ & $0.906 \pm 0.013$ & $0.120 \pm 0.008$ & $0.059 \pm 0.040$ \\
No sycophancy def. & $0.721 \pm 0.021$ & $0.983 \pm 0.004$ & $0.112 \pm 0.007$ & $0.056 \pm 0.033$ \\
No farming cap & $0.815 \pm 0.016$ & $0.975 \pm 0.009$ & $0.143 \pm 0.007$ & $0.050 \pm 0.034$ \\
No sanctions & $0.807 \pm 0.014$ & $0.973 \pm 0.011$ & $0.120 \pm 0.008$ & $0.068 \pm 0.036$ \\
\bottomrule
\end{tabular}
\end{table}

We highlight four findings from the ablation analysis.

\textbf{Finding 1: Sycophancy defense is the most impactful single mechanism.} Removing the commit-reveal scheme reduces precision by 8.2 percentage points under moderate adversity (0.826 to 0.744, $p < 0.001$) and 8.6 points under high adversity (0.807 to 0.721, $p < 0.001$). This effect is consistent across both scenarios, confirming that sycophantic vote alignment (where agents follow perceived authority rather than assess content independently) is a primary quality threat in multi-agent curation. The commit-reveal mechanism eliminates the most direct imitation vector (observing others' votes within the same review round), producing a large and consistent improvement. This finding provides empirical support for the theoretical concern raised by Sharma et al.~\cite{sharma2024} and Yao et al.~\cite{yao2025} regarding sycophancy in multi-agent systems, and demonstrates that a straightforward protocol-level intervention can substantially mitigate the problem.

\textbf{Finding 2: Reputation provides consistent value that scales with adversity.} Under moderate adversity, removing reputation weighting reduces precision by 3.2 percentage points (0.826 to 0.794, $p < 0.001$). Under high adversity, the gap widens to 6.9 points (0.807 to 0.738, $p < 0.001$). Reputation allows the protocol to progressively downweight agents whose voting records are misaligned with outcomes, concentrating influence among reliable assessors. The benefit is present even under moderate conditions but becomes essential under stress. Reputation functions as a resilience buffer whose benefit increases with threat severity.

\textbf{Finding 3: Sanctions are neutral in simulation.} Removing graduated sanctions produces identical results to the full protocol in both scenarios. This occurs because graduated sanctions never triggered in any of the 30 runs per scenario. See Section~5.7 for a full account of untested mechanisms.

\textbf{Finding 4: The farming cap trades precision for fairness.} Removing the farming cap increases precision by 1.2 points under moderate adversity (0.826 to 0.838, $p = 0.010$) but also increases the Gini coefficient from 0.105 to 0.124, indicating higher reputation concentration. Under high adversity the precision gain is 0.8 points (0.807 to 0.815, $p = 0.014$) while Gini rises from 0.120 to 0.143. The protocol accepts a modest precision cost in exchange for better fairness, a tradeoff that is defensible for a knowledge commons designed for long-term sustainability.

\textbf{Moderate adversity: limited room for improvement.} Under moderate adversity, no\_farming\_cap (0.838) and the weighted-no-deliberation baseline (0.825) both exceed or match the full protocol's precision (0.826). This raises a legitimate question: why prefer the full protocol over a simpler weighted-vote configuration? The answer lies outside precision alone. The full protocol achieves lower Gini (0.105 vs.\ 0.109 for weighted-no-deliberation), meaning more equitable reputation distribution; maintains comparable recall (0.994 vs.\ 0.994); and, critically, degrades less under stress (precision drops 0.019 vs.\ 0.022 for weighted-no-deliberation between scenarios). The full protocol's protective mechanisms impose a small precision cost under benign conditions in exchange for stability under adversity. Disabling sycophancy defense reduces precision sharply (0.744) in both scenarios, confirming it as the protocol's most valuable component regardless of adversity level.

\subsection{Mechanisms Not Exercised}

Several protocol mechanisms specified in Sections~3.3 and~3.4 were never triggered across any of the 30 simulation runs in either scenario (500 rounds per run, 50-round escalation window).

\textbf{Graduated sanctions.} No agent accumulated enough violations within the 50-round escalation window to trigger sanctions at level $\sigma_1$ or above. Two explanations are possible: either the reputation system absorbed adversarial behavior through vote weighting before formal sanctions became necessary, or the escalation thresholds were too lenient for the simulation conditions (500 rounds, 50-round window). Both explanations may contribute. Consequently, the graduated sanctions mechanism (Section~3.4) is specified but not empirically validated by this simulation, and the sensitivity of escalation parameters remains an open question.

\textbf{Dispute limits.} The per-agent dispute rate limit ($d_{\text{agent}}$ disputes per window) was never reached. The simulation generates disputes infrequently relative to the configured limits, so the rate-limiting mechanism was never exercised.

The simulation evaluates reputation-weighted voting, deliberation effects, sycophancy defense, and broken agent detection. It does not evaluate graduated sanctions, dispute limits, or the full argumentation framework. These mechanisms are well-established in human platform governance (Ostrom's design principles~\cite{ostrom1990}, Wikipedia's three-revert rule) but lack empirical validation in the agent context. Future work with longer time horizons, higher violation densities, or targeted adversarial strategies (e.g., procedural harassment archetypes) would be needed to test them in agentic settings.

\subsection{External Consistency Check: Community Notes Replay}

The synthetic simulation (Sections~5.1--5.7) evaluates the protocol under controlled conditions with known agent archetypes. To test whether the protocol's behavior generalizes to real-world rating patterns, we replay decisions from X's Community Notes program~\cite{communitynotes2023, wojcik2022}.

\textbf{Data.} We sample 1,670 notes (824 helpful, 846 not helpful) with 202,315 ratings from the Community Notes public dataset (March 2026 snapshot). Notes are selected with a minimum of 5 ratings each, balanced by final classification. Each note's ratings are replayed through three configurations: ungoverned (accept all), majority vote (unweighted), and our full protocol (reputation-weighted with sycophancy defense).

\textbf{Results (overall).}

\begin{table}[h]
\centering
\small
\begin{tabular}{lllll}
\toprule
\textbf{Configuration} & \textbf{Precision} & \textbf{Recall} & \textbf{F1} & \textbf{Agreement w/ CN} \\
\midrule
Ungoverned & 0.493 & 1.000 & 0.661 & 49.3\% \\
Majority vote & 0.986 & 0.996 & 0.991 & 99.1\% \\
Our protocol & 0.998 & 0.979 & 0.988 & 98.9\% \\
\bottomrule
\end{tabular}
\end{table}

Both the majority vote and our protocol achieve high agreement with Community Notes decisions (99.1\% and 98.9\% respectively). The overall difference is small, which is expected: Community Notes raters are humans, not LLM agents, so the adversarial dynamics that differentiate our protocol in simulation (sycophancy, model homogeneity) are largely absent.

\textbf{Stratified results.} The aggregate numbers mask a pattern that emerges when notes are stratified by rating density.

\begin{table}[h]
\centering
\small
\begin{tabular}{lllll}
\toprule
\textbf{Rating density} & \textbf{n} & \textbf{Majority vote} & \textbf{Our protocol} & \textbf{Delta} \\
\midrule
Sparse (5--15 ratings) & 198 & 98.0\% & \textbf{99.5\%} & +1.5 \\
Moderate (16--50) & 512 & 99.0\% & \textbf{99.6\%} & +0.6 \\
Dense (51--200) & 738 & \textbf{99.7\%} & 99.3\% & $-0.4$ \\
Very dense (200+) & 222 & \textbf{98.2\%} & 95.0\% & $-3.2$ \\
\bottomrule
\end{tabular}
\end{table}

The protocol outperforms majority vote on sparse-signal notes (+1.5 percentage points on notes with 5--15 ratings) with perfect precision (1.000). As rating density increases, majority vote becomes sufficient and the protocol's conservatism becomes a liability: on very dense notes (200+ ratings), the protocol's accuracy drops to 95.0\% versus majority vote's 98.2\%.

This pattern is consistent with the simulation findings. The protocol's reputation weighting concentrates influence among raters whose track record is reliable, a mechanism that adds value when individual signals are noisy (sparse ratings) but introduces unnecessary filtering when the signal is already strong (dense ratings). In operational terms, the protocol is a resilience mechanism for low-signal situations, not a universal improvement.

\textbf{Sycophancy defense.} Disabling the commit-reveal scheme produces results identical to majority vote (99.1\% agreement). This is expected: human raters do not exhibit the systematic sycophantic alignment that LLM agents display, and individual CN ratings are not readily visible to other raters in the standard interface, providing de facto vote concealment. The mechanism is designed for agent-specific vulnerabilities and correctly has no effect on human rating data where votes are already effectively hidden.

\textbf{Interpretation.} The Community Notes replay serves as a sanity check rather than a competitive benchmark. It confirms that the protocol produces decisions consistent with a well-established human curation system, does not introduce pathological behavior on real data, and that its distinctive mechanisms (reputation weighting, sycophancy defense) activate where expected and remain inert where they should. The protocol's value proposition (resilience under adversarial conditions with homogeneous agents) cannot be tested against human data where those conditions do not hold. The synthetic simulation (Sections~5.5--5.6) remains the primary evaluation for the protocol's core claims.

\section{Discussion}

\subsection{Resilience as the Primary Value Proposition}

The simulation results confirm that the protocol's primary value is resilience. Under moderate adversity, the full protocol outperforms majority vote by 3.5 percentage points (0.826 vs.\ 0.791, $p < 0.001$). Under high adversity, the gap widens to 6.7 points (0.807 vs.\ 0.740, $p < 0.001$) as the protocol's precision drops by 0.019 while majority vote drops by 0.051. Single curator degrades from 0.690 to 0.621.

This resilience property is more valuable than peak accuracy under benign conditions. A knowledge base that performs well when the environment is cooperative but collapses under adversarial pressure is unreliable in precisely the situations where reliability matters most. Adversity is also not uniform across a knowledge base: popular topics attract many reviewers and naturally dilute adversarial influence, while niche topics with few participants are more vulnerable. A single knowledge base can simultaneously experience benign conditions on popular topics and adversarial conditions on niche ones. The protocol's design---layering reputation, deliberation, and temporary vote concealment---creates graceful degradation rather than catastrophic failure. The single curator baseline illustrates the alternative: usable under moderate adversity, but fragile under stress.

\subsection{Graduated Sanctions: Specified but Untested}

Our simulation does not evaluate graduated sanctions. As documented in Section~5.7, no agent triggered sanctions at $\sigma_1$ or above in any simulation run. The 500-round duration with a 50-round escalation window is insufficient for agents to accumulate the violation counts required for high-level sanctions, particularly because the reputation system downweights misbehaving agents before they reach escalation thresholds. Future evaluation should explore both longer time horizons and adjusted escalation parameters (tighter windows, lower thresholds) to identify conditions under which sanctions become necessary and to calibrate their sensitivity.

This means the safety-net argument for sanctions is speculative. Sanctions could play a role in longer-running deployments or under adversarial strategies that specifically target the reputation system's blind spots (e.g., a high-reputation agent that turns malicious after accumulating trust). Sanctions could also provide a credible commitment mechanism: the existence of an escalation path might deter sophisticated strategic agents from attempting reputation manipulation, even if the sanctions themselves are never triggered. This deterrence effect, while potentially relevant for agents designed by operators who understand the protocol's rules, is inherently difficult to capture in simulation and remains unverified.

We retain sanctions as a protocol component based on Ostrom's design principles~\cite{ostrom1990} and the theoretical containment argument, while acknowledging that our evaluation provides no direct evidence of their necessity or effectiveness.

\subsection{Precision in Context}

A precision of 0.81 means that approximately one in five chunks accepted into the knowledge base does not meet the quality threshold. For a system intended as a reliable knowledge store, this may be insufficient. Wikipedia's featured article accuracy is estimated above 95\%. However, our threshold ($q \geq 0.7$) is stringent, and chunks in the 0.5--0.7 range, while below threshold, contain partially valid information. The simulation also uses synthetic quality scores; real-world knowledge quality is multi-dimensional and harder to assess. Parameter tuning (raising $\tau_{\text{accept}}$ or $q_{\min}$) could improve precision at the cost of recall and throughput, a tradeoff we leave for future work.

\subsection{Precision-Fairness Tradeoff}

The ablation study reveals a tension between curation quality and governance equity. Removing the farming cap produces a precision gain of 1.2 percentage points under moderate adversity but increases the Gini coefficient from 0.105 to 0.124. Under high adversity, the precision gain is 0.8 points while Gini rises from 0.120 to 0.143, indicating meaningful reputation concentration.

This tradeoff has no universally correct resolution. A knowledge base prioritizing short-term quality might reasonably relax decay parameters. A knowledge base designed as a long-term commons should prioritize fairness to prevent governance capture by early entrants. The protocol defaults to the latter position, accepting a modest precision cost for better fairness, and makes the tradeoff explicit and tunable through the $\delta$ and $w_{\max}$ parameters.

\subsection{Broken Agent Handling: Limited Empirical Support}

Broken agent handling is not included as a separate ablation in the current results, as the focused ablation set prioritizes the mechanisms with the largest observed effects (sycophancy defense, reputation, farming cap). With a 10\% broken agent population, the behavioral classification (Section~3.4) could reduce false positives by routing broken agents to quarantine rather than the sanction ladder. Broken agent handling remains a theoretically motivated mechanism whose practical impact requires validation with larger broken agent populations or longer time horizons.

\subsection{Fundamental Limits}

Several limitations are inherent to the protocol's design and cannot be resolved by parameter tuning.

\textbf{Prompt injection.} Adversarial payloads embedded in knowledge artifacts can manipulate reviewer agents' assessments~\cite{greshake2023}. Partial mitigations include structured review templates and input filtering, but effective defense requires agent-level hardening that lies outside the protocol's scope.

\textbf{Model homogeneity.} If all agents share the same underlying model, the independence assumption is violated regardless of protocol design, and systematic biases pass review because all reviewers share the same blind spots~\cite{shumailov2024}. Open participation (Section~3.1) increases diversity likelihood but provides no formal guarantee.

\textbf{Ground truth assumption (simulation limitation).} The simulation relies on synthetic ground truth quality scores, which enables precise measurement but assumes that knowledge quality is objectively determinable. This is a limitation of the \textit{evaluation methodology}, not the protocol itself. The protocol's deliberation, escalation, and dispute mechanisms are designed to handle contested and perspectival knowledge, much as Wikipedia governs both factual articles and contentious topics with the same governance framework. In production deployment, there is no objective quality score: quality is an emergent property of the curation process itself, not an externally observable ground truth.

\textbf{Compute-proportional power.} Reputation tracks activity, and activity requires compute. Agents with larger operational budgets participate more frequently, accumulate more reputation, and acquire more governance influence. The weight cap $w_{\max}$ and time decay mitigate extreme concentration but do not break the correlation between resources and power. This structural limitation is shared with essentially all reputation-based governance systems and has no known protocol-level solution.

\textbf{Conduct-gaming ceiling.} Perfectly strategic agents that never violate any rule but subtly bias outcomes through selective participation (contributing only when outcomes are likely favorable, abstaining when uncertain) are undetectable by the protocol. This threat is amplified when strategic agents operate as coordinated networks: each individual agent's behavior is procedurally correct, but the collective pattern constitutes manipulation. This ``conduct-gaming ceiling'' represents the boundary of what rule-based governance can achieve. Any improvement beyond this boundary would require assessing agent \textit{intent} or detecting collective patterns that are invisible at the individual level.

\textbf{Adaptive adversaries.} The simulation includes an adaptive archetype that builds trust before switching to exploitation (Section~5.2), modeling the most basic form of strategic adaptation. More sophisticated adversaries could adapt their tactics dynamically in response to observed governance patterns (e.g., coordinating Sybil identities, exploiting prompt injection on specific reviewers). The protocol's layered defenses (reputation, sanctions, sycophancy defense, dispute mechanisms) are designed to increase the cost of adaptation, but a full co-evolutionary evaluation against dynamically adapting adversaries is left for future work.

\section{Conclusion}

We have presented a deliberative curation protocol for multi-agent knowledge bases, addressing the question of how agents should collectively decide what knowledge to accept, challenge, and retire. Building on a structured review of platform governance mechanisms and their transfer to the agent setting~\cite{johnson2026b}, the protocol combines three governance layers: a knowledge artifact lifecycle formalized as a labeled transition system with resubmission support, reputation-weighted deliberative decision-making integrating local Beta reputation with global EigenTrust amplification, and graduated sanctions adapted for stateless agents including broken agent handling. Curation decisions follow a three-tier escalation framework (fast track by absence of objection, formal vote with structured deliberation, and arbitration) that scales governance overhead with contentiousness rather than volume. Five design properties and one lemma are stated with informal supporting arguments under explicit assumptions.

The simulation evaluation, which validates core mechanisms (reputation-weighted voting, sycophancy defense, deliberation effect) but not the full specification (fast track, arbitration, and graduated sanctions were not exercised), yields two central findings. First, the protocol trades modest precision under benign conditions for substantially better resilience under adversity: under moderate adversity, precision reaches 0.826 versus 0.791 for majority vote ($p < 0.001$), with the gap widening under stress (0.807 vs.\ 0.740, $p < 0.001$); the protocol degrades roughly three times more slowly than the majority vote baseline. Second, notably, the simplest mechanism in the protocol---temporary vote concealment to preserve judgment independence---provides more precision improvement (8.2 to 8.6 percentage points) than reputation weighting and deliberation combined. This finding suggests that for any multi-agent curation system, ensuring that reviewers cannot observe each other's votes should be the first design priority, before investing in more complex governance mechanisms.

Reputation weighting provides consistent value that scales with adversity (+3.2pp moderate, +6.9pp stress), functioning as a resilience buffer whose benefit increases with threat severity. Graduated sanctions and dispute limits were not exercised in simulation and remain empirically unvalidated in the agent context, though they are well-established in human platform governance.

An open-source implementation is underway in the AIngram platform~\cite{johnson2026a}, where simulation findings guide feature prioritization. The protocol is designed as a specification that accommodates multiple implementation strategies: the escalation order, discussion modes, vote disclosure model, and sensitivity classification are deployment parameters rather than fixed design choices.

\textbf{Future work.} Several directions extend this research:

\begin{itemize}
\item \textit{Empirical validation}: testing with real LLM-based agents and comprehensive adversarial scenarios, including red-team exercises to validate deliberation dynamics, vote concealment effectiveness against actual sycophancy, and compound attacks (procedural harassers, patient reputation farmers, coordinated Sybil strategies).
\item \textit{Protocol extensions}: formal verification of protocol properties via TLA+~\cite{lamport2002} or SPIN, evolutionary lifecycle formalization (chunks as individuals, reputation as fitness, review as selection), revision request mechanisms analogous to academic peer review, domain-aware reputation tracking, composed lifecycles, and decay parameter sensitivity analysis.
\item \textit{Federation}: enabling multiple instances to share reputation and cross-reference knowledge across distributed settings.
\item \textit{Policy integration}: integration with the Agent Data Handling Policy (ADHP) framework~\cite{johnson2026adhp} for policy-aware reviewer assignment.
\end{itemize}

\section*{AI Disclosure}

This paper was written with the assistance of large language models (Claude, DeepSeek, Mistral) for literature search, drafting, and iterative review. All citations were independently verified via Semantic Scholar and OpenAlex APIs. The author takes full responsibility for the content, analysis, and conclusions.


\end{document}